# Towards Efficient Patient Recruitment for Clinical Trials: Application of a Prompt-Based Learning Model

Mojdeh Rahmanian [a], Seyed Mostafa Fakhrahmad [a, *], Seyedeh Zahra Mousavi [b]

*[a] Department of Computer Science and Engineering and IT, Shiraz University, Shiraz, Iran*

*[b] Radiology Department, Shiraz University of Medical Sciences, Shiraz, Iran*

[*]**Corresponding author:**

Seyed Mostafa Fakhrahmad

Department of Computer Science and Engineering and IT, Shiraz University, Shiraz, Iran

Email: Fakhrahmad@shirazu.ac.ir

**Abstract**

**Objective**: Clinical trials are essential for advancing pharmaceutical interventions, but they face a bottleneck in selecting eligible participants. Although leveraging electronic health records (EHR) for recruitment has gained popularity, the complex nature of unstructured medical texts presents challenges in efficiently identifying participants. Natural Language Processing (NLP) techniques have emerged as a solution with a recent focus on transformer models. In this study, we aimed to evaluate the performance of a prompt-based large language model for the cohort selection task from unstructured medical notes collected in the EHR.

**Methods:** To process the medical records, we selected the most related sentences of the records to the eligibility criteria needed for the trial. The SNOMED CT concepts related to each eligibility criterion were collected. Medical records were also annotated with MedCAT based on the SNOMED CT ontology. Annotated sentences including concepts matched with the criteria-relevant terms were extracted. A prompt-based large language model (Generative Pre-trained Transformer (GPT) in this study) was then used with the extracted sentences as the training set. To assess its effectiveness, we evaluated the model's performance using the dataset from the 2018 n2c2 challenge, which aimed to classify medical records of 311 patients based on 13 eligibility criteria through NLP techniques.

**Results:** Our proposed model showed the overall micro and macro F measures of 0.9061 and 0.8060 which were among the highest scores achieved by the experiments performed with this dataset. Also, the current model showed the highest performance for five of the criteria among all the other ML-based published articles.

**Conclusion:** The application of a prompt-based large language model in this study to classify patients based on eligibility criteria received promising scores. Besides, we proposed a method of extractive summarization with the aid of SNOMED CT ontology that can be also applied to other medical texts.

1. **Introduction**

Achieving sufficient research participant enrolment is essential to conduct a successful clinical trial. These participants should fulfill a list of eligibility criteria based on the goal of the trial. The shortage of systematic health records for each patient and the required manual reviewing and selecting eligible cases have made cohort selection the most cumbersome and time-consuming step of clinical trials (1).

In recent years, the development of electronic health records (EHR) has improved this step to some extent. However, the utilization of unstructured or semi-structured ambiguous text in EHR has limited its application. Moreover, the absence of standardized terminology and diverse documentation styles complicates this issue. To address the above challenges, researchers have increasingly turned to Natural Language Processing (NLP) techniques (2). Until now, different rule-based and machine learning (ML)-based NLP models have been investigated in biomedical domains, with the most recent experiments related to transformers (3-6). Transformers have emerged as a revolutionary advancement in the field of NLP in recent years, fundamentally reshaping the way computers understand and generate human language (7-14).

Recently, large language models (LLMs) such as GPT-3 and GPT-4 have achieved notable progress in the field of NLP. Given the computational challenges of fine-tuning these massive models, researchers explored prompt-based learning as an alternative. Prompts can be manually designed using few-shot learning, where a few task examples are provided, or zero-shot learning, where the model is instructed on the expected output type.

Motivated by these promising results, we aim to investigate the performance of a prompt-based model for the cohort selection tasks. This paper presents the application of prompt design in the context of text classification and explores its efficacy in a specific domain-focused classification task. The main focus is investigating whether adopting a prompt-based learning strategy can yield outcomes comparable to alternative techniques. To assess this, we employ the cohort identification of patients meeting eligibility criteria from narrative medical records, as part of the track 1 shared task in the 2018 National NLP Clinical Challenges (n2c2). The challenge provided free-text medical records of 311 patients consisting of different. Also, 13 eligibility criteria required for clinical trials were listed and the system should provide decisions regarding the eligibility of each patient for these 13 separate criteria.

2. **Review of Articles:**

In recent years, different rule-based and machine learning (ML)-based NLP models have been investigated in biomedical domains, with the most recent experiments related to transformers (15-24). Xiong et al. used trigger words followed by machine reading comprehension (MRC) including BiDAF (bi-directional attention flow), BiMPM (bilateral multi-perspective matching), and BERT frameworks. Among these models, NCBI-BERT with attention mechanism received the highest micro F score (5).

Cahyawijay et al. investigated the performance of pre-trained LMs such as general BERT models and biomedical and clinical BERT models including clinicalBERT, PubMedBERT, BioELECTRA, and bio-lm. In general, clinical domain BERT models outperformed the general domain models (6).

The field of natural language processing is experiencing a revolution due to the rapid advancements in Large Language Models (LLMs). The application of LLMs in the medical domain has received considerable attention in recent times (25), with diverse applications including clinical documentation (26, 27), clinical decision support (28, 29), knowledge-based medical information retrieval and generation (30-32), and medical research (33-38). Additionally, LLMs can address the challenge of insufficient medical text data for training clinical NLP models, aiding in data augmentation and performing tasks such as data collection, processing, and analysis (39, 40).

However, the application of LLMs for clinical trial matching is in the early stages (41). One of the rare examples of this topic is related to the 2021 and 2022 Clinical Trial track. In these challenges, inclusion/exclusion criteria of a set of clinical trials were retrieved from a registration platform (i.e. ClinicalTrials.gov). Besides, some patient case descriptions were provided and the participants should label each patient as excluded, included, or not relevant for each retrieved clinical trial (42). Among the published results of the participants of these tracks, only Jin et al. applied a large language model for this purpose (43). However, patient notes used in this study showed a mean length of up to 9 sentences and 154 words. Therefore, these short texts cannot evaluate the efficacy of LLMs accurately. Also, most of the retrieved clinical trials didn't include complex inclusion/exclusion criteria which need complicated reasoning tasks. Therefore, we selected the 2018 n2c2 data for our investigation since it provided more cases with significantly longer free-text data for each case associated with more diverse and complex trial criteria.

3. **Materials and Methods**

*3.1.   Dataset*

We utilized the dataset from the 2018 n2c2 shared task (track 1), which focused on cohort selection for clinical trials. The dataset comprised 311 patient records, manually labeled by experts to indicate whether a patient meets one of 13 possible eligibility criteria. The detailed description of these criteria is summarized in Table 1. The dataset was divided into training (202 records) and test (86 records) sets. Figure 1 illustrates the distribution of the 13 criteria in both datasets, revealing significant imbalances. The test dataset exhibits a similar criteria distribution to the training dataset.

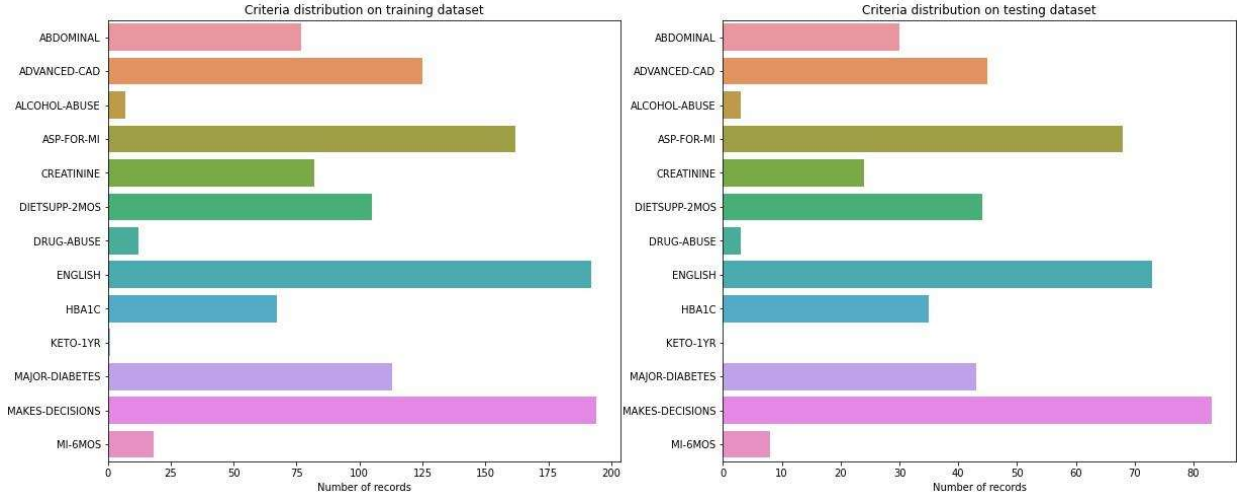

**Figure 1.** Distribution of the 13 eligibility criteria in the training and test datasets provided in the 2018 n2c2 shared task track 1. The number of the records labeled as "met" for each criterion is illustrated in the figure for the training and testing sets, separately.

### 3.2. Methodology

Figure 2 presents an overview of the main stages in the proposed framework. In the following paragraphs, a comprehensive breakdown of each step will be provided.

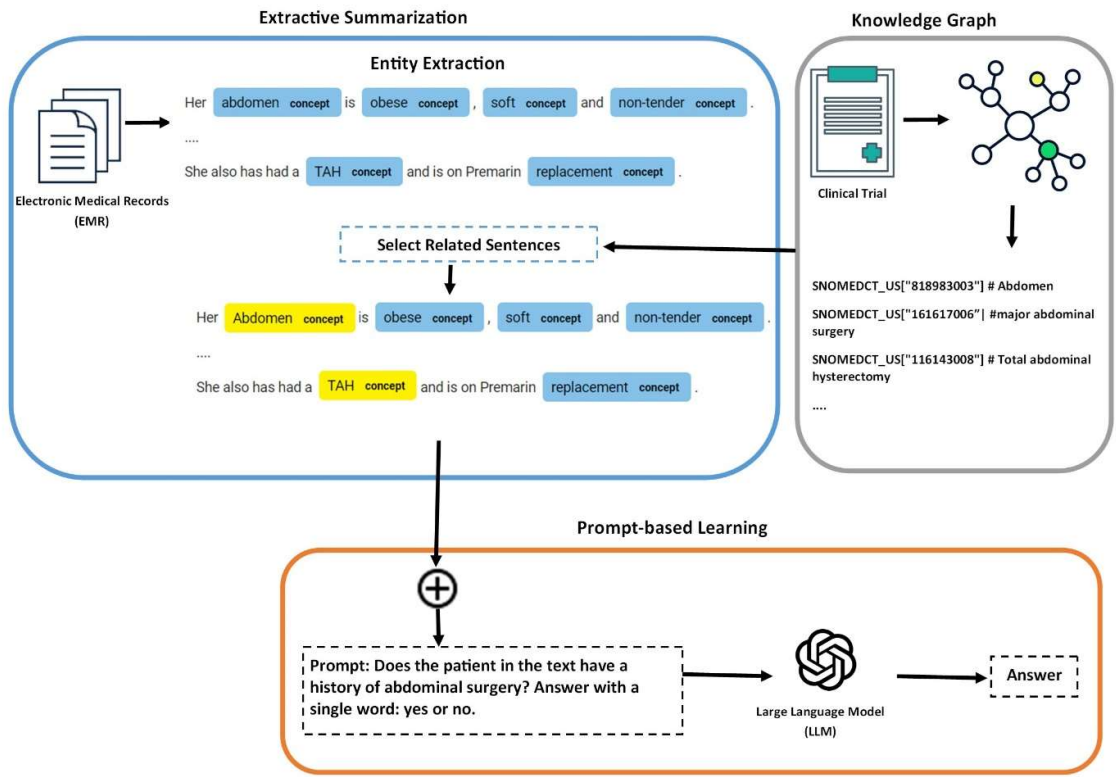

**Figure 2.** A summary of the proposed framework for the application of prompt-based learning for cohort selection.

### 3.2.1 Knowledge Graph

In contrast to other related studies where trigger words were manually collected from various sources for text processing, our proposed method involves leveraging the concepts and relations encoded in SNOMED CT within the UMLS to extract information from eligibility criteria and each patient record, analogous to the approach used in (18) for measuring the similarity between the eligibility criteria and patients. For instance, the eligibility criterion: "ABDOMINAL: History of intra-abdominal surgery, small or large intestine resection, or small bowel obstruction." can be converted into several SNOMED CT keywords including "Major abdominal surgery", "Small intestine excision", "Large intestine excision", and "Small bowel obstruction". Then, all the descendants of these key terms were retrieved automatically with their concept unique identifier (CUI) codes by employing an ontology-based approach. To facilitate this process, we utilized the PyMedTermino2 (44) Python package, which enabled us to map CUI codes to SNOMED CT codes. We constructed a code list for each criterion comprising the primary keywords; for example, "161617006, major abdominal surgery [Finding]" along with its associated descendants.

### 3.2.2 Extractive Summarization

In this stage, clinical free texts were automatically annotated by MedCAT (45), an NLP tool that builds upon sciSpaCy to recognize named entities associated with clinical concepts. These entities were subsequently linked to the standard terminologies, namely SNOMED CT and UMLS. Our approach utilized a publicly available MedCAT model, which had been pre-trained on the complete text data within the MIMIC-III database and the entirety of SNOMED CT. In the final step, temporal information and sentences of the records encompassing annotated concepts with a CUI code included within the code list created for each criterion were extracted.

### 3.2.3 Prompt Engineering

In the prompt engineering step, we converted each criterion to a prompt and gave the prompt with the summaries produced in the previous step to the GPT model. A prompt, in the context of language models, is an instruction or question that is given to the model to guide its response. The challenge defined each criterion and we converted these definitions to questions or prompts. Table 1 illustrates the questions constructed for the 13 criteria. In our approach to binary classification using the GPT-3 model, we have developed a straightforward method centered on the utilization of question-based prompts limited to one-word responses: either "yes" or "no." To transform this response into a binary classification result, we establish a mapping between the model's answer and interpret "yes" as the class "met" and "no" as the class "not met". We instruct the model to perform the identification of each criterion for every summarized electronic medical record. We evaluate our prompt against all 86 records in the test set. An illustrative example of zero-shot learning for recognizing eligibility criteria: "ABDOMINAL" is presented in Figure 3.

**Table 1.** 13 included eligibility criteria with their definitions and the created question-based prompts

| Cohort Selection Criteria | Definition provided by the challenge | Question-based Prompt |
|---|---|---|
| ABDOMINAL | History of intra-abdominal surgery, small or large intestine resection, or small bowel obstruction | Does the patient in the following text have a history of abdominal surgery? Answer with one word yes or no. |
| ADVANCED-CAD | Advanced cardiovascular disease (CAD). For this annotation, we define "advanced" as having 2 or more of the following: Taking 2 or more medications to treat CAD, history of myocardial infarction (MI), currently experiencing angina, and ischemia, past or present. | Does the patient her/himself in the following text have two or more of the four following criteria? Taking 2 or more medications to treat CAD, history of myocardial infarction (MI), currently experiencing angina, and ischemia |
| ALCOHOL-ABUSE | Current alcohol use over weekly recommended limits | Is the patient in the following text currently using alcohol over the weekly recommended limits? |
| ASP-FOR-MI | Use of aspirin to prevent MI | Is the patient of the following text using aspirin to prevent myocardial infarction (mi) and no other diseases? |
| CREATININE | Serum creatinine > upper limit of normal | Does the patient of the following text have creatinine or cr level of larger than 1.4? |
| DIETSUPP-2MOS | Taken a dietary supplement (excluding vitamin D) in the past 2 months | Please review the patient's record and determine if any of the dietary supplements were mentioned in the record date labeled (current time) or a maximum of two months before that. If so, answer yes otherwise no. Consider each record date mentioned at the beginning of each paragraph as YYYY-MM-DD. Dietary supplements such as: Folic acid, Multivitamins, Vitamins excluding vitamin D, calcium, magnesium, iron, Echinacea and ginger, Caffeine and curcumin, Tryptophan and glutamine, Probiotics, Glucosamine, and Fish oils. |
| DRUG-ABUSE | Drug abuse, current or past | Is the patient in the following text experiencing drug abuse? Answer with one word yes or no. |
| ENGLISH | The patient must speak English | Is the patient speaking a language other than English in the text below? Answer with one word yes or no. |
| HBA1C | Any hemoglobin A1c (HbA1c) value between 6.5% and 9.5% | Has the patient ever had a hemoglobin value between 6.5 and 9.5? If at least one time is mentioned, just answer 'yes' otherwise 'no'. |
| KETO-1YR | Diagnosis of ketoacidosis in the past year | Has the patient in the following text been diagnosed with ketoacidosis in the past year? Consider the record date labeled (current time) as a current time and each record date mentioned at the beginning of each paragraph as YYYY-MM-DD. |
| MAJOR-DIABETES | Major diabetes-related complications. For this annotation, we define "major complications" (as opposed to "minor complications") as any of the following that are a result of (or strongly correlated with) uncontrolled diabetes: amputation, kidney damage, skin conditions, retinopathy, nephropathy, and neuropathy | Does the patient of the following text have an amputation or disease related to kidney or skin or retinopathy or nephropathy or neuropathy? If at least one time is mentioned, just answer 'yes' otherwise 'no'. |
| MAKES-DECISIONS | Patients must make their own medical decisions | Does the patient in the following text have cognitive limitations and cannot make their own medical decisions? Answer with yes or no. |

| MI-6MOS | MI in the past 6 months | Does the patient in the following text have any myocardial infarction mentioned in the Record date labeled (current time) or a maximum of 6 months before that, considering the date mentioned at the beginning of each paragraph in the format of record date: YYYY-MM-DD as the hole paragraph's correct date -not the dates in the text. |
|---|---|---|

---

**Prompt:** Does the patient in the following text have a history of abdominal surgery? Answer with one word yes or no. The text is delimited with triple backticks.

text: "' Bladder cancer status post resection in 2089.Status post hysterectomy in 2065. R adrenal mass--dx on ab CT 2094 L adnexal cystic lesion Colon polyps s/p scope with resection Diverticulosis Medications: CVS: RRR nl s1 s2. no m/r/g Abdomen: +BS, soft, minimally tender in lower quadrants, no organomegaly, guaiac negative.'"

**Answer:** Yes.

**Figure 3.** An example of a prompt-based approach for one of the criteria.

### 3.2.4 Evaluation Metrics

We used the evaluation method provided by the challenge. Results of the GPT were given to a predefined evaluation model created by the challenge to receive the precision, recall, F score, specificity, and AUC metrics. For our experiment, precision shows how many of the records labeled as "met" are correct (true "met"). Recall means how many of the "met" records in the dataset are correctly labeled in our model. Specificity is a measure of how many records labeled as "not met" are correct. F score is a metric calculated by an equation combining both precision and recall. There are micro and macro F scores with some differences in the equation. It is shown that the micro F score is a better predictor of the model when there is imbalanced data.

## 4. Experiment and Results

### 4.1. Model

For our research, we utilized OpenAI's powerful GPT-3.5 Turbo model, recognized as OpenAI's most capable and cost-effective GPT-3.5 variant (46). To ensure the reproducibility of our results across multiple iterations, we consistently set the temperature to 0 for all requests. Although configuring the temperature to 0 significantly enhances the determinism of the outputs, there may still be a slight degree of variability.

*4.2. Results*

To demonstrate the impact of the summarization method, we presented the results in tables separately for two scenarios: once without using the summarization method and once using the summarization method. In our first scenario, we didn't perform any summarization. However, given the model's text processing constraints which limit the input length to approximately 4096 tokens, meaning the maximum number of words (or semantically related segments) that can be provided as input is restricted to around 4096 words, we decided to truncate the medical texts and excluded the conclusions of clinical narratives. It's worth noting that only about 11% of the records exceed a length of 4000 words. Table 2 provides an overview of the findings in the first scenario. We measured the overall micro and macro F scores and the precision, recall, and AUC metrics for each criterion, separately.

**Table 2.** A summary of the evaluation metrics overall and for each criterion, separately without using the summarization method.

|  | Met | | | | Not met | | | Overall | |
| --- | --- | --- | --- | --- | --- | --- | --- | --- | --- |
|  | Precision | Recall | Specificity | F (b=1) | Precision | Recall | F (b=1) | F (b=1) | AUC |
| ABDOMINAL | 0.5455 | 0.2000 | 0.9107 | 0.2927 | 0.6800 | 0.9107 | 0.7786 | 0.5357 | 0.5554 |
| ADVANCED-CAD | 0.8148 | 0.4889 | 0.8780 | 0.6111 | 0.6102 | 0.8780 | 0.7200 | 0.6656 | 0.6835 |
| ALCOHOL-ABUSE | 0.0000 | 0.0000 | 0.9880 | 0.0000 | 0.9647 | 0.9880 | 0.9762 | 0.4881 | 0.4940 |
| ASP-FOR-MI | 1.0000 | 0.0735 | 1.0000 | 0.1370 | 0.2222 | 1.0000 | 0.3636 | 0.2503 | 0.5368 |
| CREATININE | 0.8571 | 0.2500 | 0.9839 | 0.3871 | 0.7722 | 0.9839 | 0.8652 | 0.6262 | 0.6169 |
| DIETSUPP-2MOS | 1.0000 | 0.0455 | 1.0000 | 0.0870 | 0.5000 | 1.0000 | 0.6667 | 0.3768 | 0.5227 |
| DRUG-ABUSE | 1.0000 | 0.3333 | 1.000 | 0.5000 | 0.9765 | 1.0000 | 0.9881 | 0.7440 | 0.6667 |
| ENGLISH | 0.8488 | 1.0000 | 0.0000 | 0.9182 | 0.0000 | 0.0000 | 0.0000 | 0.4591 | 0.5000 |
| HBA1C | 0.0000 | 0.0000 | 1.0000 | 0.0000 | 0.5930 | 1.0000 | 0.7445 | 0.3723 | 0.5000 |
| KETO-1YR | 0.0000 | 0.0000 | 1.0000 | 0.0000 | 1.0000 | 1.0000 | 1.0000 | 0.5000 | 0.5000 |
| MAJOR-DIABETES | 1.0000 | 0.3488 | 1.0000 | 0.5172 | 0.6056 | 1.0000 | 0.7544 | 0.6358 | 0.6744 |
| MAKES-DECISIONS | 0.9765 | 1.0000 | 0.3333 | 0.9881 | 1.0000 | 0.3333 | 0.5000 | 0.7440 | 0.6667 |
| MI-6MOS | 0.0000 | 0.0000 | 0.9487 | 0.0000 | 0.9024 | 0.9487 | 0.9250 | 0.4625 | 0.4744 |
| Overall (micro) | 0.8730 | 0.4641 | 0.9530 | 0.6060 | 0.7185 | 0.9530 | 0.8193 | 0.7126 | 0.7085 |
| Overall (macro) | 0.6187 | 0.2877 | 0.8494 | 0.3414 | 0.6790 | 0.8494 | 0.7140 | 0.5277 | 0.5686 |

After employing our proposed summarization method in the second scenario, we observed a notable improvement in the model's performance. By strategically summarizing the clinical narratives, we effectively reduced the input length while preserving key information. Table 3 summarizes the results of our experiment using the summarization. We measured the overall micro and macro F scores and the precision, recall, F score, and AUC metrics for each criterion, separately. Our proposed model received the final micro and macro F measures of 0.9035 and 0.7949, respectively. ALCOHOL-ABUSE, CREATININE, ENGLISH, and ABDOMINAL achieved the highest scores with F measures above 0.9 while KETO-1YR, MI-6MOS, and MAKES-DECISIONS were the lowest with F scores below 0.7.

**Table 3.** A summary of the evaluation metrics overall and for each criterion, separately using the SNOMED CT-based summarization method.

|  | Met | | | | Not met | | | Overall | |
| --- | --- | --- | --- | --- | --- | --- | --- | --- | --- |
|  | Precision | Recall | Specificity | F (b=1) | Precision | Recall | F (b=1) | F (b=1) | AUC |
| ABDOMINAL | 0.9000 | 0.9000 | 0.9464 | 0.9000 | 0.9464 | 0.9464 | 0.9464 | 0.9232 | 0.9232 |
| ADVANCED-CAD | 0.7843 | 0.8889 | 0.7317 | 0.8333 | 0.8571 | 0.7317 | 0.7895 | 0.8114 | 0.8103 |
| ALCOHOL-ABUSE | 1.0000 | 1.0000 | 1.0000 | 1.0000 | 1.0000 | 1.0000 | 1.0000 | 1.0000 | 1.0000 |
| ASP-FOR-MI | 0.9041 | 0.9706 | 0.6111 | 0.9362 | 0.8462 | 0.6111 | 0.7097 | 0.8229 | 0.7908 |
| CREATININE | 1.0000 | 0.8750 | 1.0000 | 0.9333 | 0.9538 | 1.0000 | 0.9764 | 0.9549 | 0.9375 |
| DIETSUPP-2MOS | 0.9677 | 0.6818 | 0.9762 | 0.8000 | 0.7455 | 0.9762 | 0.8454 | 0.8227 | 0.8290 |
| DRUG-ABUSE | 1.0000 | 0.3333 | 1.0000 | 0.5000 | 0.9765 | 1.0000 | 0.9881 | 0.7440 | 0.6667 |
| ENGLISH | 0.9730 | 0.9863 | 0.8462 | 0.9796 | 0.9167 | 0.8462 | 0.8800 | 0.9298 | 0.9162 |
| HBA1C | 0.8077 | 0.6000 | 0.9020 | 0.6885 | 0.7667 | 0.9020 | 0.8288 | 0.7587 | 0.7510 |
| KETO-1YR | 0.0000 | 0.0000 | 0.9884 | 0.0000 | 1.0000 | 0.9884 | 0.9942 | 0.4971 | 0.4942 |
| MAJOR-DIABETES | 0.9412 | 0.7442 | 0.9535 | 0.8312 | 0.7885 | 0.9535 | 0.8632 | 0.8472 | 0.8488 |
| MAKES-DECISIONS | 0.9762 | 0.9880 | 0.3333 | 0.9820 | 0.5000 | 0.3333 | 0.4000 | 0.6910 | 0.6606 |
| MI-6MOS | 0.3636 | 0.5000 | 0.9103 | 0.4211 | 0.9467 | 0.9103 | 0.9281 | 0.6746 | 0.7051 |
| Overall (micro) | 0.9068 | 0.8693 | 0.9378 | 0.8877 | 0.9115 | 0.9378 | 0.9245 | 0.9061 | 0.9035 |
| Overall (macro) | 0.8168 | 0.7283 | 0.8615 | 0.7542 | 0.8649 | 0.8615 | 0.8577 | 0.8060 | 0.7949 |

Table 4 compares the scores of our model with the best results of other previously mentioned ML-based experiments performed in this field with this n2c2 dataset on average and for each criterion separately. Promisingly, our model achieved the second rank regarding the micro F measure and the highest macro F score. Moreover, the model received the highest F scores for five of the thirteen criteria including ABDOMINAL, CREATININE, ASP-FOR-MI, ENGLISH, and ALCOHOL-ABUSE.

**Table 4.** Comparison of the results of our proposed method with the best results of other ML-based experiments published with this dataset.

|  | The current study | Hassanzadeh et al. (enriched multi-layer perception) | Segura-Bedmar et al. (hybrid CNN+RNN approach) | Xiong, Peng et al. (NCBI-BERT + attention) | Chen et al. (medical knowledge-infused CNN) | Spasic et al. | Xiong, Shi et al. (LSTM-highway-LSTM) |
|---|---|---|---|---|---|---|---|
| ABDOMINAL | 0.9232 | 0.7226 | 0.4792 | 0.8530 | 0.7680 | 0.7677 | 0.7811 |
| ADVANCED-CAD | 0.8114 | 0.7654 | 0.3478 | 0.8947 | 0.6840 | 0.8814 | 0.8103 |
| ALCOHOL-ABUSE | 1.0000 | 0.4911 | 0.4915 | 0.4911 | 0.7440 | 0.6417 | 0.4911 |
| ASP-FOR-MI | 0.8229 | 0.4416 | 0.4340 | 0.7735 | 0.7430 | 0.7442 | 0.7734 |
| CREATININE | 0.9549 | 0.7480 | 0.5581 | 0.7955 | 0.7180 | 0.8716 | 0.8380 |
| DIETSUPP-2MOS | 0.8227 | 0.7185 | 0.6703 | 0.8586 | 0.7560 | 0.8350 | 0.8371 |
| DRUG-ABUSE | 0.7440 | 0.4911 | 0.4828 | 0.7441 | 0.7560 | 0.7378 | 0.6910 |
| ENGLISH | 0.9298 | 0.4591 | 0.4737 | 0.8303 | 0.7920 | 0.7929 | 0.8745 |
| HBA1C | 0.7587 | 0.6098 | 0.4000 | 0.8259 | 0.7730 | 0.9253 | 0.9253 |
| KETO-1YR | 0.4971 | 0.5000 | 0.5000 | 0.5000 | 0.5000 | 0.5000 | 0.5000 |
| MAJOR-DIABETES | 0.8472 | 0.8372 | 0.4665 | 0.8954 | 0.8140 | 0.8254 | 0.8602 |
| MAKES-DECISIONS | 0.6910 | 0.4911 | 0.4828 | 0.7440 | 0.4910 | 0.6910 | 0.7440 |
| MI-6MOS | 0.6746 | 0.4756 | 0.4828 | 0.4756 | 0.4760 | 0.6605 | 0.7933 |
| Overall micro F score | 0.9061 | 0.8399 | 0.7856 | 0.9070 | 0.8610 | 0.8904 | 0.9021 |
| Overall macro F score | 0.8060 | 0.5962 | 0.4823 | - | 0.6730 | - | - |

5.      Discussion

Utilization of a prompt-based learning model associated with large language models such as GPT to facilitate automatic cohort selection in this article showed promising results. To better clarify the process, the results of one of the criteria that is "ALCOHOL-ABUSE" were summarized in the supplementary material. Supplementary file 1 showed the list of the SNOMED CT terms retrieved for the "ALCOHOL-ABUSE". Supplementary file 2 presented the selected sentences for each test record that contained one of the terms listed in supplementary file 1 associated with the correct label and the label provided by GPT to compare.  Label 0 corresponds to "not met" and label 1 means "met".

Our model received the overall micro and macro F scores of 0.9061 and 0.8060 with acceptable performance for all criteria. To evaluate the performance of our model for each criterion separately, we classified the criteria based on the NLP method required for each of them as mentioned by Stubbs et al. (4):

- Concept extraction

Four criteria including ABDOMINAL, MAJOR-DIABETES, CREATININE, and HBA1C were related to this category requiring finding and extracting some clinical terms and numerical values.

To facilitate this process, we used the related SNOMED CT concepts for each criterion. Although some of the previous articles have used the theory of criteria-relevant trigger terms to shorten the records and select the most relevant parts to extract critical concepts, many of them have manually created specific dictionaries for each criterion based on various sources with the manual intervention of a medical expert based on the review of the training dataset that lacked reproducibility and reliability

For the two numerical criteria, the main reason for the false-negative results was writing the concepts in short forms and abbreviations. Since these abbreviations were not included in the SNOMED CT ontology, the model cannot identify and select them. For example, "creatinine" was written as "Cr" or "Creat" in some records and parsing data formatted unusually was difficult for the model to learn. This problem can be solved through pre-processing steps. However, we didn't apply pre-processing techniques to have a better assessment of the performance of the model when faced with medical narratives with different writing styles.

- Temporal reasoning

Four criteria including DIETSUPP-2MOS, MI-6MOS, ADVANCED-CAD, and KETO-1YR needed temporal processing besides extracting the relevant concepts. Since each clinical data consisted of medical records written at different time points and the mentioned criteria required the analysis of the dates, we extracted all the record dates. Therefore, the selected relevant sentences were classified based on their time point.

Regarding DIETSUPP-2MOS, the main advantage of our article was to use SNOMED CT ontology to get the names of the associated supplements the same as all the criteria. Since searching and retrieving the various diet supplements from the UMLS and in particular SNOMED CT ontology was a challenging process, most of the experiments had manually created a list of supplements with the aid of a medical expert or available websites. However, the main limitation of our study was the commercial names and abbreviations of the supplements.

For MI-6MOS, although our system received the second-highest F score among the available ML-based experiments, the main limitation of our study was false negative results due to the incompleteness of the relevant SNOMED CT terms. Some records included information such as the history of previous angioplasty or coronary artery bypass graft (CABG) to emphasize the patient history of MI without a direct reference to the terms related to the infarction. However, this issue can be easily solved by adding other SNOMED CT concepts related to these missed terms.

Regarding KETO-1YR, our model received almost the same score as the previously proposed methods due to the imbalance of the met and not-met cases with only one positive training record which makes it difficult for ML-based approaches to learn.

For ADVANCED-CAD, the patient should present with two or more of four signs or symptoms. The results showed the correct selection of the related sentences and proper identification of time points required for these criteria. However, the exact definition of these required signs and symptoms was not determined. For instance, some antihypertensive drugs are also prescribed for patients with ischemic heart disease even without definite signs of hypertension and it is not clear whether these medications fulfill the criteria or not.

- Inference

ASP-FOR-MI, MAKES-DECISIONS, and ENGLISH needed reasoning; therefore, we expected the ML-based models to show a better result. Our model received the highest score among the other ML-based experiments for the ASP-FOR-MI and ENGLISH; however, MAKES-DECISIONS was more challenging. Evaluating the mislabelled records revealed that the main limitation was also related to the concept extraction, not the reasoning process. Since it relates to a broad clinical pathology with a variety of signs and symptoms, extracting the most relevant SNOMED CT concepts was cumbersome. Adding all the concepts related to neurological, mental, and cognitive pathologies in the annotation process will solve this problem to some extent.

About DRUG-ABUSE and ALCOHOL-ABUSE, our model also showed promising results with a few false negative results due to the lack of the uncommon names of some drugs in the SNOMED CT concept.

Moreover, the strengths of GPT models for cohort selection will be more evident when larger datasets are included due to some reasons. First, in clinical practice where providing labeled training data is cumbersome, GPT models will be promising solutions since they don't require task-specific training data. Second, GPT models are the best choices when a language model with a large knowledge base is required since they are pre-trained extensively on various text sources. In electronic health records, we are facing medical notes related to different topics written in various styles and formats. Therefore, a large-scale language model can be more practical when more data from different centers are included. Third, as we mentioned in the article, we performed minimal preprocessing on the data before running the model. Preprocessing is one of the time-consuming and essential steps of NLP models requiring expertise and different processing applications. Achieving an acceptable result with a GPT model in our study without the need to perform the preprocessing step will increase the speed of the clinical matching task while reducing the need for an expert to perform the preprocessing (47-49).

Besides the mentioned points, we introduced a method of extractive summarization with the help of SNOMED CT ontology and an annotator tool to capture the most crucial information from the original text. To evaluate the performance of the summarization method, some of the medical records included in the challenge dataset were given to the question-based prompts before and after the summarization. It was revealed that the use of the original texts showed the wrong answers for some of the records while the application of the summarized texts solved the problem and gave the correct labels.

**Declaration of Competing Interest:**
The authors declare that they have no known competing financial interests or personal relationships that could have appeared to influence the work reported in this paper.

**Funding:**
This research did not receive any specific grant from funding agencies in the public, commercial, or not-for-profit sectors.

**CRediT authorship contribution statement:**
**Mojdeh Rahmanian:** Conceptualization, Formal Analysis, Methodology, Software, Writing-Original Draft, Writing-Review & Editing.  **Seyed Mostafa Fakhrahmad:** Project Administration, Resources, Supervision, Review & Editing.  **Seyedeh Zahra Mousavi:** Conceptualization, Methodology, Writing-Original Draft, Writing-Review & Editing.
**Availability of data and materials:** The data used in this study was part of track 1 shared task in the 2018 National NLP Clinical Challenges (n2c2) and is accessible online through the challenge website.